\documentclass[journal]{IEEEtran}

\usepackage{url}
\usepackage{latexsym}

\usepackage{times}
\usepackage{amsmath}
\usepackage{amsthm}
\usepackage{amssymb}
\usepackage{graphicx}
\usepackage{xspace}
\usepackage{tabularx}
\usepackage{multicol}
\usepackage{multirow}
\usepackage{url}
\usepackage{wrapfig}
\usepackage{comment}
\usepackage{listings}
\usepackage{color}
\usepackage[utf8]{inputenc}
\usepackage{fancyvrb}
\usepackage{booktabs}
\usepackage{color}
\usepackage[normalem]{ulem}

\newcommand{\bmx}[0]{\begin{bmatrix}}
\newcommand{\emx}[0]{\end{bmatrix}}

\newcommand{\vect}[1]{\mathbf{#1}}

\newcommand{\matr}[1]{\mathbf{#1}}

\newcommand{\vc}[0]{\vect{c}}

\newcommand{\vh}[0]{\vect{h}}
\newcommand{\vv}[0]{\vect{v}}
\newcommand{\vx}[0]{\vect{x}}
\newcommand{\vz}[0]{\vect{z}}
\newcommand{\vw}[0]{\vect{w}}

\newcommand{\vf}[0]{\vect{f}}

\newcommand{\vy}[0]{\vect{y}}
\newcommand{\vg}[0]{\vect{g}}

\newcommand{\vu}[0]{\vect{u}}

\newcommand{\vr}[0]{\vect{r}}
\newcommand{\mW}[0]{\matr{W}}

\newcommand{\mU}[0]{\matr{U}}

\newcommand{\LL}[0]{\mathcal{L}}

\newcommand{\RR}[0]{\mathbb{R}}

\newcommand{\ola}{\overleftarrow}
\newcommand{\ora}{\overrightarrow}

\begin{document}
%
\title{Describing Multimedia Content using Attention-based Encoder--Decoder Networks}
%
%
%

\author{Kyunghyun Cho$^\circ$, Aaron Courville$^\circ$ and Yoshua Bengio$^{\circ
    \star}$
    \thanks{$^\circ$ Universit\'e de Montr\'eal 
            $^\star$ CIFAR Senior Fellow}
}

\maketitle

\begin{abstract}
Whereas deep neural networks were first mostly used for classification tasks,
they are rapidly expanding in the realm of structured output problems, where the
observed target is composed of multiple random variables that have a rich joint
distribution, given the input. We focus in this paper on the case where the
input also has a rich structure and the input and output structures are somehow
related. We describe systems that learn to {\em attend} to different places in
the input, for each element of the output, for a variety of tasks: machine
translation, image caption generation, video clip description and speech
recognition. All these systems are based on a shared set of building blocks:
gated recurrent neural networks and convolutional neural networks, along with
trained {\em attention mechanisms}. We report on experimental results with these
systems, showing impressively good performance and the advantage of the
attention mechanism.
\end{abstract}


%
\IEEEpeerreviewmaketitle

\section{Introduction}

\IEEEPARstart{I}{n} this paper we focus on the application of deep
learning to structured output problems where the task is to map the input to an
output that possesses its own structure. The task is therefore not only to map
the input to the correct output (e.g. the classification task in object recognition), but also to
model the structure within the output sequence. 

A classic example of a
structured output problem is machine translation: to automatically translate a
sentence from the \emph{source language} to the \emph{target language}. To
accomplish this task, not only does the system need to be concerned with
capturing the semantic content of the source language sentence, but also with
forming a coherent and grammatical sentence in the target language. In other words, given
an input source sentence, we cannot choose the elements of the output
(i.e. the individual words) independently: they have a complex joint distribution.

Structured output problems represent a large and important class of problems
that include classic tasks such as speech recognition and many natural language
processing problems (e.g. text summarization and paraphrase generation). As the
range of capabilities of deep learning systems increases, less established forms
of structured output problems, such as image caption generation and video
description generation (\cite{Kulkarni2013} and references therein,) are being
considered. 

One important aspect of virtually all structured output tasks is that {\em the
structure of the output is imtimately related to the structure of the
input}. A central challenge to these tasks is therefore the problem of
\emph{alignment}. At its most fundamental, the problem of alignment is the
problem of how to relate sub-elements of the input to sub-elements of the
output. Consider again our example of machine translation. In order to
translate the source sentence into the target language we need to first
decompose the source sentence into its constituent semantic parts. Then we
need to map these semantic parts to their counterparts in the target
language. Finally, we need to use these semantic parts to compose the
sentence following the grammatical regularities of the target
language. Each word or phrase of the target sentence can be aligned to a
word or phrase in the source language.

In the case of image caption generation, it is often appropriate for the output sentence to accurately describe 
the spatial relationships between elements of the scene represented in the image. For this, we need to \emph{align} the output words to spatial
regions of the source image.

In this paper we focus on a general approach to the alignment problem known
as the soft attention mechanism. Broadly, attention mechanisms are components of
prediction systems that allow the system to sequentially focus on different
subsets of the input. The selection of the subset is typically conditioned
on the state of the system which is itself a function of the previously
attended subsets.

Attention mechanisms are employed for two purposes. The first is to reduce
the computational burden of processing high dimensional inputs by selecting
to only process subsets of the input. The second is to allow the system to
focus on distinct aspects of the input and thus improve its ability to
extract the most relevant information for each piece of the output, thus
yielding improvements in the quality of the generated outputs.

As the name suggests, soft attention mechanisms avoid a hard selection of
which subsets of the input to attend and instead uses a soft weighting of the
different subsets. Since all subset are processed,
these mechanisms offer no computation advantage. Instead, the advantage brought
by the soft-weighting is that it is readily amenable to efficient learning via
gradient backpropagation.

In this paper, we present a review of the recent work in applying the soft 
attention to structured output tasks and spectulate about the future course of this line of research. The soft-attention mechanism is part of a growing litterature on more flexible deep learning architectures that embed a certain amount of \emph{distributed decision making}.

\section{Background: \\ Recurrent and Convolutional Neural Networks}

\subsection{Recurrent Neural Network}
\label{sec:rnn}

A recurrent neural network (RNN) is a neural network specialized at handling a
variable-length input sequence $x=(\vx_1, \dots, \vx_T)$ and optionally a
corresponding variable-length output sequence $y=(\vy_1, \dots, \vy_T)$, using
an internal hidden state $\vh$. The RNN sequentially reads each symbol $\vx_t$
of the input sequence and updates its internal hidden state $\vh_t$ according to
\begin{align}
    \label{eq:encoding}
    \vh_t = \phi_{\theta}\left( \vh_{t-1}, \vx_t \right),
\end{align}
where $\phi_{\theta}$ is a nonlinear activation function parametrized by a set of
parameters $\theta$. When the target sequence is given, the RNN can be trained to
sequentially make a prediction $\hat{\vy}_t$ of the actual output $\vy_t$ at each time step $t$:
\begin{align}
    \label{eq:seq_pred}
    \hat{\vy}_t = g_{\theta}\left( \vh_{t}, \vx_t \right),
\end{align}
where $g_{\theta}$ may be an arbitrary, parametric function that is learned
jointly as a part of the whole network.

The recurrent activation function $\phi$ in Eq.~\eqref{eq:encoding} may be as
simple as an affine transformation followed by an element-wise logistic function
such that
\begin{align*}
    \vh_t = \tanh \left( \mU \vh_{t-1} + \mW \vx_t \right),
\end{align*}
where $\mU$ and $\mW$ are the learned weight matrices.\footnote{
    We omit biases to make the equations less cluttered.
}

It has recently become more common to use more sophisticated recurrent
activation functions, such as a long short-term memory
(LSTM,~\cite{Hochreiter+Schmidhuber-1997}) or a gated recurrent unit
(GRU,~\cite{Cho-et-al-EMNLP2014,Chung-et-al-NIPSDL2014-small}), to reduce the
issue of vanishing gradient~\cite{Bengio-et-al-TNN1994,Hochreiter+al-2000}.
Both LSTM and GRU avoid the vanishing gradient by introducing gating units that
adaptively control the flow of information across time steps. 

The activation of a GRU, for instance, is defined by
\begin{align*}
    \vh_t = \vu_t \odot \tilde{\vh}_t + (1 - \vu_t) \odot \vh_{t-1},
\end{align*}
where $\odot$ is an element-wise multiplication, and the update gates $\vu_t$
are 
\begin{align*}
    \vg_t = \sigma\left( \mU_u \vh_{t-1} + \mW_u \vx_t \right).
\end{align*}
The candidate hidden state $\tilde{\vh}_t$ is computed by
\begin{align*}
    \tilde{\vh}_t = \tanh\left( \mU \vh_{t-1} + \mW \left(\vr_t \odot \vx_t\right) \right),
\end{align*}
where the reset gates $\vr_t$ are computed by
\begin{align*}
    \vr_t = \sigma\left( \mU_r \vh_{t-1} + \mW_r \vx_t \right).
\end{align*}

All the use cases of the RNN in the remaining of this paper use either the GRU
or LSTM.

\subsection{RNN-LM: Recurrent Neural Network Language Modeling}
\label{sec:rnnlm}

In the task of language modeling, we let a model learn the probability
distribution over natural language sentences. In other words, given a model, we
can compute the probability of a sentence $s=(w_1, w_2, \ldots, w_T)$ consisting
of multiple words, i.e., $p(w_1, w_2, \ldots, w_T)$, where the sentence is $T$
words long. 

This task of language modeling is equivalent to the task of predicting the next
word. This is clear by rewriting the sentence probability into
\begin{align}
    \label{eq:next_word_pred}
    p(w_1, w_2, \ldots, w_T) = \prod_{t=1}^T p(w_t \mid w_{<t}),
\end{align}
where $w_{<t}=(w_1, \ldots, w_{t-1})$.  Each conditional probability on the
right-hand side corresponds to the predictive probability of the next word $w_t$
given all the preceding words ($w_1, \ldots, w_{t-1}$). 

A recurrent neural network (RNN) can, thus, be readily used for language
modeling by letting it predict the next symbol at each time step $t$
(RNN-LM, \cite{Mikolov-ICASSP-2011}). In other words, the RNN predicts the probability
over the next word by
\begin{align}
    \label{eq:rnn_lm}
    p(w_{t+1} = w|w_{\leq t}) = g_{\theta}^w \left( \vh_{t}, \vw_t
    \right),
\end{align}
where $g_{\theta}^w$ returns the probability of the word $w$ out of all possible
words. The internal hidden state $\vh_{t}$ summarizes all the preceding symbols
$w_{\leq t} = (w_1,\ldots,w_{t})$.  

We can generate an exact sentence sample from an RNN-LM by iteratively sampling
from the next word distribution $p(w_{t+1}|w_{\leq t})$ in
Eq.~\eqref{eq:rnn_lm}. Instead of stochastic sampling, it is possible to
approximately find a sentence sample that maximizes the probability $p(s)$
using, for instance, beam search~\cite{Graves2012,Boulanger2013}.

The RNN-LM described here can be extended to learn a {\it conditional} language
model. In conditional language modeling, the task is to model the distribution
over sentences given an additional input, or context. The context may be
anything from an image and a video clip to a sentence in another language. Examples
of textual outputs associated with these inputs by the {\it conditional RNN-LM} include 
respectively an image caption, a video description and a translation. In these cases,
the transition function of the RNN will take as an additional input the context
$c$ such that
\begin{align}
    \label{eq:encoding_cond}
    \vh_t = \phi_{\theta}\left( \vh_{t-1}, \vx_t , c\right).
\end{align}
Note the $c$ at the end of the r.h.s. of the equation.

This conditional language model based on RNNs will be at the center of later
sections.

\subsection{Deep Convolutional Network}
\label{sec:cnn}

A convolutional neural network (CNN) is a special type of a more general
feedforward neural network, or multilayer perceptron, that has been specifically
designed to work well with two-dimensional images \cite{LeCun+98}. The CNN often
consists of multiple convolutional layers followed by a few fully-connected
layers.

At each convolutional layer, the input image of width $n_i$, height $n_j$ and $c$
color channels ($\vx \in \RR^{n_i \times n_y \times c}$) is first convolved with a
set of local filters $\vf \in \RR^{n'_i \times n'_y \times c \times d}$. For each
location/pixel $(i,j)$ of $\vx$, we get
\begin{align}
    \label{eq:conv}
    \vz_{i,j} = \sum_{i'=1}^{n'_i} \sum_{j'=1}^{n'_j} f\left(\vf_{i',j'}^{\top}
    \vx_{i+i',j+j'}\right), 
\end{align}
where $\vf_{i',j'} \in \RR^{c \times d}$, $\vx_{i+i',j+j'} \in \RR^{c}$ and
$\vz_{i,j} \in \RR^{d}$. $f$ is an element-wise nonlinear activation function.

The convolution in Eq.~\eqref{eq:conv} is followed by local max-pooling:
\begin{align}
    \label{eq:maxpool}
    \vh_{i,j} = \max_{
        \begin{array}{c}
            i'\in\left\{ ri, \ldots, (r+1)i - 1\right\}, \\
            j'\in\left\{ rj, \ldots, (r+1)j - 1\right\}
        \end{array}
    } \vz_{i',j'},
\end{align}
for all $i\in\left\{ 1, \ldots, n_i/r \right\}$ and $j\in\left\{ 1, \ldots, n_j/r
\right\}$. $r$ is the size of the neighborhood. 

The pooling operation has two desirable properties. First, it reduces the
dimensionality of a high-dimensional output of the convolutional layer.
Furthermore, this spatial max-pooling summarizes the activation of the
neighbouring feature activations, leading to the (local) translation invariance.

After a small number of convolutional layers, the final feature map from the
last convolutional layer is flattened to form a vector representation $\vh$ of
the input image. This vector $\vh$ is further fed through a small number of
fully-connected nonlinear layers until the output.

Recently, the CNNs have been found to be excellent at the task of large-scale
object recognition. For instance, the annual ImageNet Large Scale Visual
Recognition Challenge (ILSVRC) has a classification track where more than a
million annotated images with 1,000 classes are provided as a training set.  In
this challenge, the CNN-based entries have been dominant since
2012~\cite{Krizhevsky-2012,ZeilerFergus14,Simonyan2015,Szegedy-et-al-arxiv2014}.

\subsection{Transfer Learning with Deep Convolutional Network}
\label{sec:imagenet}

Once a deep CNN is trained on a large training set such that the one provided as
a part of the ILVRC challenge, we can use any intermediate representation, such
as the feature map from any convolutional layer or the vector representation
from any subsequent fully-connected layers, of the whole network for tasks other
than the original classification. 

It has been observed that the use of these intermediate representation from the
deep CNN as an image descriptor significantly boosts subsequent tasks such as
object localization, object detection, fine-grained recognition, attribute
detection and image retrieval (see, e.g., \cite{Sermanet14,razavian2014cnn}.)
Furthermore, more non-trivial tasks, such as image caption generation
\cite{Karpathy+Li-CVPR2015,Fang-et-al-CVPR2015,Mao-et-al-ICLR2015,Vinyals-et-al-CVPR2015,Kiros-et-al-arxiv2014},
have been found to benefit from using the image descriptors from a pre-trained
deep CNN. In later sections, we will discuss in more detail how image
representations from a pre-trained deep CNN can be used in these non-trivial
tasks such as image caption generation~\cite{Xu-et-al-ICML2015} and video
description generation~\cite{Li-et-al-ARXIV2015}.

\section{Attention-based Multimedia Description}

Multimedia description generation is a general task in which a model generates a
natural language description of a multimedia input such as speech, image and
video as well as text in another language, if we take a more general view. This
requires a model to capture the underlying, complex mapping between the
spatio-temporal structures of the input and the complicated linguistic
structures in the output. In this section, we describe a neural network based
approach to this problem, based on the encoder--decoder framework with the
recently proposed attention mechanism.

\subsection{Encoder--Decoder Network}
\label{sec:encdec}

An encoder--decoder framework is a general framework based on neural networks that
aims at handling the mapping between highly structured input and output. It was
proposed recently in
\cite{Kalchbrenner+Blunsom-EMNLP2013,Cho-et-al-EMNLP2014,Sutskever-et-al-NIPS2014}
in the context of machine translation, where the input and output are natural
language sentences written in two different languages.

As the name suggests, a neural network based on this encoder--decoder framework
consists of an encoder and a decoder. The encoder $f_{\text{enc}}$ first reads
the input data $x$ into a continuous-space representation $c$: 
\begin{align}
    \label{eq:enc_gen}
    c = f_{\text{enc}}(x),
\end{align}

The choice of $f_{\text{enc}}$ largely depends on the type of input. When $x$ is
a two-dimensional image, a convolutional neural network (CNN) from
Sec.~\ref{sec:imagenet} may be used. A recurrent neural network (RNN) in
Sec.~\ref{sec:rnn} is a natural choice when $x$ is a sentence.

The decoder then generates the output $y$ conditioned on the continuous-space
representation, or context $c$ of the input. This is equivalent to computing the
conditional probability distribution of $y$ given $x$:
\begin{align}
    \label{eq:dec_gen}
    p(Y|x) = f_{\text{dec}}(c).
\end{align}
Again, the choice of $f_{\text{dec}}$ is made based on the type of the output.
For instance, if $y$ is an image or a pixel-wise image segmentation, a conditional
restricted Boltzmann machine (CRBM) can be used \cite{taylor-eccv-10-small}.
When $y$ is a natural language description of the input $x$, it is natural to
use an RNN which is able to model natural languages, as described in
Sec.~\ref{sec:rnnlm}.

\begin{figure}[ht]
    \centering
    \includegraphics[width=0.65\columnwidth]{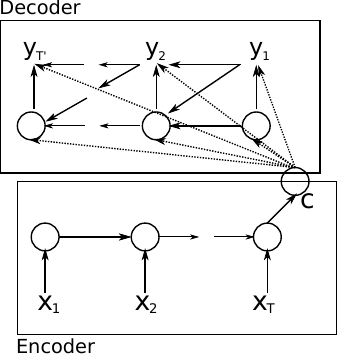}
    \caption{Graphical illustration of the simplest form encoder-decoder model for machine
    translation from \cite{Cho-et-al-EMNLP2014}. $x=(x_1, \ldots, x_T)$,
$y=(y_1, \ldots, y_{T'})$ and $c$ are respectively the input sentence, the
output sentence and the continuous-space representation of the input sentence.}
    \label{fig:encdec_mt}
\end{figure}

This encoder--decoder framework has been successfully used in
\cite{Sutskever-et-al-NIPS2014,Cho-et-al-EMNLP2014} for machine translation. In
both work, an RNN was used as an encoder to summarize a source sentence (where
the summary is the last hidden state $\vh_T$ in Eq.~\eqref{eq:encoding}) from
which a conditional RNN-LM from Sec.~\ref{sec:rnn} decoded out the corresponding
translation. See Fig.~\ref{fig:encdec_mt} for the graphical illustration.

In \cite{Mao-et-al-ICLR2015,Vinyals-et-al-CVPR2015}, the authors used a
pre-trained CNN as an encoder and a conditional RNN as a decoder to let model
generate a natural language caption of images. Similarly, a simpler
feedforward log-bilinear language model~\cite{Mnih+Hinton-2007} was used as a
decoder in \cite{Kiros-et-al-arxiv2014}. The authors of
\cite{Venugopalan-et-al-arxiv2014} applied the encoder--decoder framework to
video description generation, where they used a pre-trained CNN to extract a
feature vector from each frame of an input video and averaged those vectors.  

In all these recent applications of the encoder--decoder framework, the
continuous-space representation $c$ of the input $x$ returned by an encoder, in
Eq.~\eqref{eq:enc_gen} has been a fixed-dimensional vector, regardless of the
size of the input.\footnote{
    Note that in the case of machine translation and video description
    generation, the size of the input varies.
} Furthermore, the context vector was not structured by design, but rather an
arbitrary vector, which means that there is no guarantee that the context vector
preserves the spatial, temporal or spatio-temporal structures of the input.
Henceforth, we refer to an encoder--decoder based model with a fixed-dimensional
context vector as a {\it simple encoder--decoder model}.

\subsection{Incorporating an Attention Mechanism}
\label{sec:attention}

\subsubsection{Motivation}

A naive implementation of the encoder--decoder framework, as in the simple
encoder--decoder model, requires the encoder to compress the input into a single
vector of predefined dimensionality, regardless of the size of or the amount of
information in the input. For instance, the recurrent neural network (RNN) based
encoder used in \cite{Cho-et-al-EMNLP2014,Sutskever-et-al-NIPS2014} for machine
translation needs to be able to summarize a variable-length source sentence into
a single fixed-dimensional vector. Even when the size of the input is fixed, as
in the case of a fixed-resolution image, the amount of information contained in
each image may vary significantly (consider a varying number of objects in each
image).

In \cite{Cho2014a}, it was observed that the performance of the neural machine
translation system based on a simple encoder--decoder model rapidly degraded as
the length of the source sentence grew. The authors of \cite{Cho2014a}
hypothesized that it was due to the limited capacity of the simple
encoder--decoder's fixed-dimensional context vector.

Furthermore, the interpretability of the simple encoder--decoder is extremely
low. As all the information required for the decoder to generate the output is
compressed in a context vector without any presupposed structure, such structure
is not available to techniques designed to inspect the representations captured by the
model~\cite{ZeilerFergus14,Springenberg2015,Denil2014c}.

\subsubsection{Attention Mechanism for Encoder--Decoder Models}
\label{sec:attention:details}

We the introduction of an attention mechanism in between the encoder and
decoder, we address these two issues, i.e., (1) limited capacity of a
fixed-dimensional context vector and (2) lack of interpretability.

The first step into introducing the attention mechanism to the encoder--decoder
framework is to let the encoder return a structured representation of the input.
We achieve this by allowing the continuous-space representation to be a set of
fixed-size vectors, to which we refer as a {\it context set}, i.e., 
\[ 
    c=\left\{\vc_1, \vc_2, \ldots, \vc_M \right\}
\]
See Eq.~\eqref{eq:enc_gen}. Each vector in the context set is localized to a
certain spatial, temporal or spatio-temporal component of the input. For instance, in the
case of an image input, each context vector $\vc_i$ will summarize a certain
spatial location of the image (see Sec.~\ref{sec:capgen}), and with machine
translation, each context vector will summarize a phrase centered around a
specific word in a source sentence (see Sec.~\ref{sec:nmt}.) In all cases, the
number of vectors $M$ in the context set $c$ may vary across input examples.

The choice of the encoder and of the kind of context set it will return is governed
by the application and the type of the input considered.
In this paper, we assume that the decoder is a conditional RNN-LM from
Sec.~\ref{sec:rnnlm}, i.e., the goal is to describe the input in a natural
language sentence.

The attention mechanism controls the input actually seen by the decoder and requires
another neural network, to which refer as the attention model. The main job of
the attention model is to score each context vector $\vc_i$ with respect to the
current hidden state $\vz_{t-1}$ of the decoder:\footnote{
    We use $\vz_t$ to denote the hidden state of the decoder to distinguish it
    from the encoder's hidden state for which we used $\vh_t$ in
    Eq.~\eqref{eq:encoding}.
} 
\begin{align}
    \label{eq:att_score_gen}
    e_i^t = f_{\text{ATT}}(\vz_{t-1}, \vc_i, \{ \alpha_j^{t-1}\}_{j=1}^M ),
\end{align}
where $\alpha_j^{t-1}$ represents the attention weights computed
at the previous time step, from the scores $e_i^{t-1}$, through
a softmax that makes them sum to 1:
\begin{align}
    \label{eq:att2}
    \alpha_i^t = \frac{\exp(e_i^t)}{\sum_{j=1}^M \exp(e_j^t)},
\end{align}
This type of scoring can be
viewed as assigning a probability of being {\it attended} by the decoder to
each context, hence the name of the attention model.

Once the attention weights are computed, we use them to compute the new context
{\it vector} $\vc^t$:
\begin{align}
    \label{eq:att_gen}
    \vc^t = \varphi\left(\left\{ \vc_i \right\}_{i=1}^M, 
        \left\{ \alpha_i^t \right\}_{i=1}^M\right),
\end{align}
where $\varphi$ returns a vector summarizing the whole context set $c$ according
to the attention weights.

A usual choice for $\varphi$ is a simple weighted sum of the context vectors
such that
\begin{align}
    \label{eq:att_wsum}
    \vc^t = \varphi\left(\left\{ \vc_i \right\}_{i=1}^M, 
    \left\{ \alpha_i^t \right\}_{i=1}^M\right) = \sum_{i=1}^M \alpha_i \vc_i.
\end{align}
On the other hand, we can also force the attention model to make a hard decision
on which context vector to consider by sampling one of the context vectors
following a categorical (or multinoulli) distribution:
\begin{align}
    \label{eq:att_hard}
    \vc^t = \vc_{r^t}, \text{ where }
    r^t \sim \text{Cat}(M, \left\{ \alpha_i^t \right\}_{i=1}^M). 
\end{align}

With the newly computed context vector $\vc_t$, we can update the hidden state
of the decoder, which is a conditional RNN-LM here, by
\begin{align}
    \label{eq:encoding_cond_att}
    \vh_t = \phi_{\theta}\left( \vh_{t-1}, \vx_t , \vc_t\right).
\end{align}

This way of computing a context vector at each time step $t$ of the decoder
frees the encoder from compressing any variable-length input into a single
fixed-dimensional vector. By spatially or temporally dividing the
input\footnote{
    Note that it is possible, or even desirable to use overlapping regions.
}, the encoder can represent the input into a set of vectors of which each
needs to encode a fixed amount of information focused around a particular
region of the input. In other words, the introduction of the attention
mechanism bypasses the issue of limited capacity of a fixed-dimensional context
vectors.

\begin{figure}[t]
    \centering
    \includegraphics[width=0.85\columnwidth]{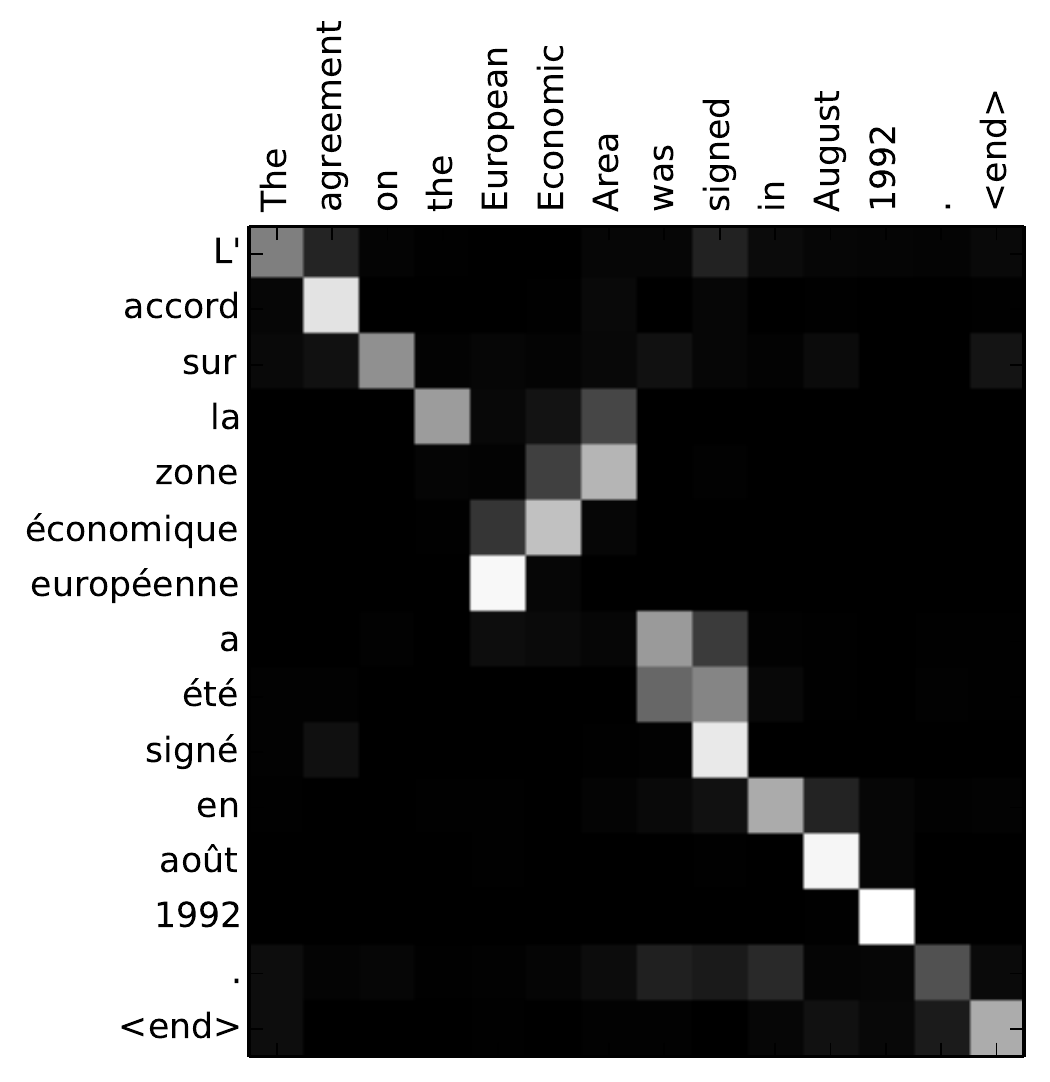}
    \caption{Visualization of the attention weights $\alpha_j^t$ of the
    attention-based neural machine translation model \cite{Bahdanau-et-al-ICLR2015-small}. Each row corresponds to
the output symbol, and each column the input symbol. Brighter the higher
$\alpha_j^t$.}
    \label{fig:alpha_viz}
\end{figure}

Furthermore, this attention mechanism allows us to directly inspect the
internal working of the whole encoder--decoder model. The magnitude of the
attention weight $\alpha_j^t$, which is positive by construction in
Eq.~\eqref{eq:att2}, highly correlates with how predictive the spatial,
temporal or spatio-temporal region of the input, to which the $j$-th context
vector corresponds, is for the prediction associated with the $t$-th output
variable $y_t$. This can be easily done by visualizing the attention matrix
$\left[ \alpha_j^t \right]_{t,j} \in \RR^{T' \times M}$, as in
Fig.~\ref{fig:alpha_viz}.

This attention-based approach with the weighted sum of the context vectors (see
Eq.~\eqref{eq:att_wsum}) was originally proposed in
\cite{Bahdanau-et-al-ICLR2015-small} in the context of machine translation,
however, with a simplified (content-based) scoring function:
\begin{align}
    \label{eq:att1}
    e_i^t = f_{\text{ATT}}(\vz_{t-1}, \vc_i).
\end{align}
See the missing $\{ \alpha_j^{t-1}\}_{j=1}^M$ from Eq.~\eqref{eq:att_score_gen}.
In \cite{Xu-et-al-ICML2015}, it was further extended with the hard attention
using Eq.~\eqref{eq:att_hard}.  In~\cite{Chorowski2015} this attention mechanism
was extended to be by taking intou account the past values of the attention
weights as the general scoring function from Eq.~\eqref{eq:att_score_gen},
following an approach based purely on those weights introduced
by~\cite{Graves-arxiv2013}. We will discuss more in detail these three
applications/approaches in the later sections.

\subsection{Learning}

As usual with many machine learning models, the attention-based encoder--decoder
model is also trained to maximize the log-likelihood of a given training set
with respect to the parameters, where the log-likelihood is defined as
\begin{align}
    \label{eq:loglikelihood}
    \LL\left(D=\left\{ (x^n, y^n) \right\}_{n=1}^N, \Theta\right) = 
    \frac{1}{N} \sum_{n=1}^N \log p(y^n \mid x^n, \Theta),
\end{align}
where $\Theta$ is a set of all the trainable parameters of the model.

\subsubsection{Maximum Likelihood Learning}
\label{sec:mle}

When the weighted sum is used to compute the context vector, as in
Eq.~\eqref{eq:att_wsum}, the whole attention-based encoder--decoder model
becomes one large differentiable function. This allows us to compute the
gradient of the log-likelihood in Eq.~\eqref{eq:loglikelihood} using
backpropagation~\cite{Rumelhart86b}. With the computed gradient, we can use, for
instance, the stochastic gradient descent (SGD) algorithm to iteratively update the
parameters $\Theta$ to maximize the log-likelihood.

\subsubsection{Variational Learning for Hard Attention Model}
\label{sec:vari}

When the attention model makes a hard decision each time as in
Eq.~\eqref{eq:att_hard}, the derivatives through the stochastic decision
are zero, because those decisions are discrete. Hence, the information
about how to improve the way to take those focus-of-attention decisions is
not available from back-propagation, while it is needed to train the
attention mechanism. The question of training neural networks
with stochastic discrete-valued hidden units has a long history,
starting with Boltzmann machines~\cite{Hinton-bo86}, with recent
work studying how to deal with such units in a system trained using back-propagated
gradients \cite{bengio2013estimating,Tang+Salakhutdinov-NIPS2013,ba+mnih,Raiko2015}.
Here we briefly describe the variational learning approach from
\cite{ba+mnih,Xu-et-al-ICML2015}.

With stochastic variables $\vr$ involved in the computation from
inputs to outputs, the log-likelihood in
Eq.~\eqref{eq:loglikelihood} is re-written into
\begin{align*}
    \LL\left(D=\left\{ (x^n, y^n) \right\}_{n=1}^N, \Theta\right) = 
    \frac{1}{N} \sum_{n=1}^N l(y^n, x^n, \Theta),
\end{align*}
where
\begin{align*}
    l(y, x, \Theta) = \log \sum_{\vr} p(y, \vr | x, \Theta)
\end{align*}
and $\vr=(r_1, r_2, \ldots, r_T')$.  We derive a lowerbound of $l$ as
\begin{align}
    l(y, x) &= \log \sum_{\vr} p(y| \vr, x) p(\vr|x) 
    \nonumber \\ 
    &\geq 
    \sum_{\vr} p(\vr|x) \log p(y| \vr, x).
    \label{eq:lowerbound}
\end{align}
Note that we omitted $\Theta$ to make the equation less cluttered.

The gradient of $l$ with respect to $\Theta$ is then
\begin{align}
    \nabla l(y, x) = \sum_{\vr} p(\vr|x) &\left[ \nabla \log p(y | \vr, x) \right.
    \nonumber
    \\
                    &+ \left. \log p (y| \vr,x) \nabla \log p(\vr|x) \right]
    \label{eq:grad_lb}
\end{align}
which is often approximated by Monte Carlo sampling:
\begin{align}
    \label{eq:approx_grad_lb}
    \nabla l(y, x) \approx \frac{1}{M} \sum_{m=1}^M &\nabla \log p(y | \vr^m, x) 
    \nonumber
    \\
    &+\log p (y| \vr^m,x) \nabla \log p(\vr^m|x). 
\end{align}
As the variance of this estimator is high, a number of variance reduction
techniques, such as baselines and variance normalization, are often used in
practice~\cite{Mnih+Gregor-2014arxiv,ba+mnih}.

Once the gradient is estimated, any usual gradient-based iterative optimization
algorithm can be used to approximately maximize the log-likelihood.

\section{Applications}

In this section, we introduce some of the recent work in which the
attention-based encoder--decoder model was applied to various multimedia
description generation tasks.

\subsection{Neural Machine Translation}
\label{sec:nmt}

Machine translation is a task in which a sentence in one language (source) is
translated into a corresponding sentence in another language (target).
{\em Neural} machine translation aims at solving it with a single neural network based
model, jointly trained end-to-end. The encoder--decoder framework described in
Sec.~\ref{sec:encdec} was proposed for neural machine translation recently in
\cite{Kalchbrenner+Blunsom-EMNLP2013,Cho-et-al-EMNLP2014,Sutskever-et-al-NIPS2014}.
Based on these works, in \cite{Bahdanau-et-al-ICLR2015-small}, the
attention-based model  was proposed to make neural machine translation systems
more robust to long sentences. Here, we briefly describe the model from
\cite{Bahdanau-et-al-ICLR2015-small}.

\subsubsection{Model Description}

The attention-based neural machine translation in
\cite{Bahdanau-et-al-ICLR2015-small} uses a bidirectional recurrent neural
network (BiRNN) as an encoder. The forward network reads the input sentence
\mbox{$x=(x_1, \ldots, x_T)$} from the first word to the last, resulting in 
a sequence of state vectors
\[
    \left\{ \ora{\vh}_1, \ora{\vh}_2, \ldots, \ora{\vh}_T \right\}.
\]
The backward network, on the other hand, reads the input sentence in the reverse
order, resulting in 
\[
    \left\{ \ola{\vh}_T, \ola{\vh}_{T-1}, \ldots, \ola{\vh}_1 \right\}.
\]
These vectors are concatenated per step to form a context set (see
Sec.~\ref{sec:attention:details}) such that $\vc_t = \left[ \ora{\vh}_t;
\ola{\vh}_t \right]$.

\begin{figure}[ht]
    \centering
    \begin{minipage}{0.45\columnwidth}
        \centering
        \includegraphics[width=\columnwidth]{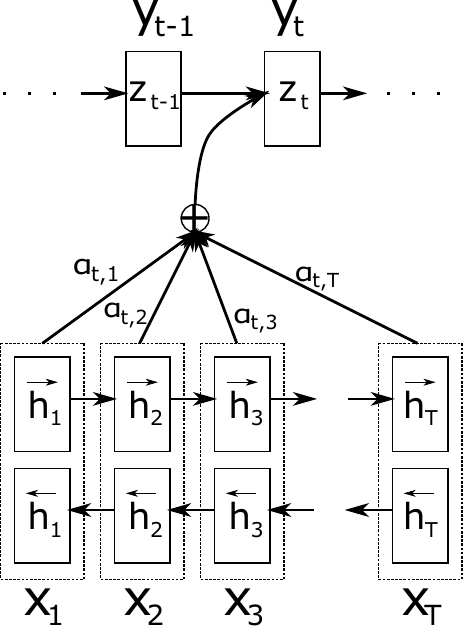}
    \end{minipage}
    \hfill
    \begin{minipage}{0.47\columnwidth}
        \caption{Illustration of a single step of decoding in
            attention-based neural machine translation
        \cite{Bahdanau-et-al-ICLR2015-small}.}
        \label{fig:nmt}
    \end{minipage}
\end{figure}

The use of the BiRNN is crucial if the content-based attention mechanism is
used. The content-based attention mechanism in
Eqs.~\eqref{eq:att1}~and~\eqref{eq:att2} relies solely on a so-called {\it
content-based scoring}, and without the context information from the whole
sentence, words that appear multiple times in a source sentence cannot be
distinguished by the attention model.

The decoder is a conditional RNN-LM that models the target language given the
context set from above. See Fig.~\ref{fig:nmt} for the graphical illustration of
the attention-based neural machine translation model.

\begin{table}[ht]
    \centering
    \caption{The translation performances and the relative improvements over the
    simple encoder-decoder model on an English-to-French translation task,
measured by BLEU \cite{Bahdanau-et-al-ICLR2015-small,Jean-et-al-ACL2015}.
$\star$: an ensemble of multiple attention-based models. $\circ$: the
state-of-the-art phrase-based statistical machine translation system~\cite{Durrani2014}.}
    \label{tbl:att_nmt}
    \begin{tabular}{c|c|c}
    Model & BLEU & Rel. Improvement \\
    \hline
    \hline
    Simple Enc--Dec & 17.82 &  -- \\
    Attention-based Enc--Dec & 28.45 & +59.7\% \\
    Attention-based Enc--Dec (LV) & 34.11 & +90.7\% \\
    Attention-based Enc--Dec (LV)$^\star$ & {\bf 37.19} & {\bf +106.0\%} \\
    \hline
    State-of-the-art SMT$^\circ$ & 37.03 & -- 
    \end{tabular}
\end{table}

\subsubsection{Experimental Result}

Given a fixed model size, the attention-based model proposed in
\cite{Bahdanau-et-al-ICLR2015-small} was able to achieve a relative
improvement of more than 50\% in the case of the English-to-French translation
task, as shown in Table~\ref{tbl:att_nmt}. When the very same model was extended with
a very large target vocabulary~\cite{Jean-et-al-ACL2015}, the relative
improvement over the baseline without the attention mechanism was 90\%.
Additionally, the very same model was recently tested on a number of European
language pairs at the WMT'15 Translation
Task.\footnote{\url{http://www.statmt.org/wmt15/}}.  See Table~\ref{tbl:wmt15}
for the results.

\begin{table}
    \centering
    \caption{The performance of the attention-based neural machine translation
            models with the very large target vocabulary in the WMT'15 Translation
            Track \cite{Jean-et-al-ACL2015}. We show the results on two
            representative language pairs. For the complete result, see
        http://matrix.statmt.org/.}
        \label{tbl:wmt15}
    \begin{tabular}{c | c c c }
        Language Pair & Model & BLEU & Note \\
        \hline
        \hline
        \multirow{2}{*}{En-$>$De} & NMT & 24.8 & \\
                                & Best Non-NMT & 24.0 & Syntactic SMT (Edinburgh) \\
        \hline
        \multirow{2}{*}{En-$>$Cz} & NMT & 18.3 & \\
                                & Best Non-NMT & 18.2 & Phrase SMT (JHU) \\
    \end{tabular}
\end{table}

The authors of \cite{Gulcehre-Orhan-et-al-2015} recently proposed a method for
incorporating a monolingual language model into the attention-based neural
machine translation system. With this method, the attention-based model was
shown to outperform the existing statistical machine translation systems on
Chinese-to-English (restricted domains) and Turkish-to-English translation
tasks as well as other European languages they tested.

\subsection{Image Caption Generation}
\label{sec:capgen}

Image caption generation is a task in which a model looks at an input image and
generates a corresponding natural language description. The encoder--decoder
framework fits well with this task. The encoder will extract the
continuous-space representation, or the context, of an input image, for
instance, with a deep convolutional network (see Sec.~\ref{sec:cnn},) and from
this representation the conditional RNN-LM based decoder generates a natural
language description of the image. Very recently (Dec 2014), a number of
research groups
independently proposed to use the simple encoder--decoder model to solve the
image caption generation
\cite{Fang-et-al-CVPR2015,Karpathy+Li-CVPR2015,Mao-et-al-ICLR2015,Vinyals-et-al-CVPR2015}.
Instead, here we describe a more recently proposed approach based on the
attention-based encoder--decoder framework in \cite{Xu-et-al-ICML2015}.

\begin{figure}[ht]
    \centering
    \includegraphics[width=0.8\columnwidth]{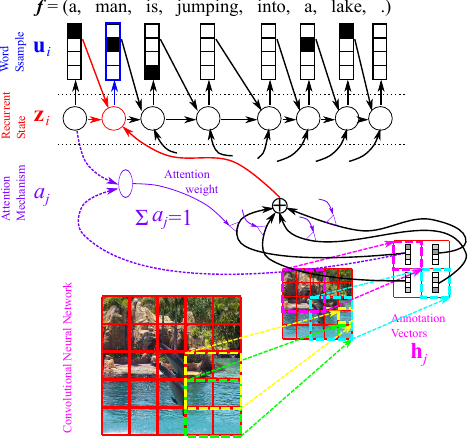}
    \caption{Graphical illustration of the attention-based encoder--decoder
    model for image caption generation.}
    \label{fig:capgen}
\end{figure}

\subsubsection{Model Description}

The usual encoder--decoder based image caption generation models use the
activation of the last fully-connected hidden layer as the continuous-space
representation, or the context vector, of the input image (see
Sec.~\ref{sec:imagenet}.) The authors of \cite{Xu-et-al-ICML2015} however
proposed to use the activation from the last {\it convolutional} layer of the
pre-trained convolutional network, as in the bottom half of
Fig.~\ref{fig:capgen}. 

Unlike the fully-connected layer, in this case, the context set consists of
multiple vectors that correspond to different spatial regions of the input image
on which the attention mechanism can be applied. Furthermore, due to convolution
and pooling, the spatial locations in pixel space represented by each context vector overlaps
substantially with those represented by the neighbouring context vectors, which
helps the attention mechanism distinguish similar objects in an image using its
context information with respect to the whole image, or the neighbouring pixels.

Similarly to the attention-based neural machine translation in
Sec.~\ref{sec:nmt}, the decoder is implemented as a conditional RNN-LM. In
\cite{Xu-et-al-ICML2015}, the content-based attention mechanism (see
Eq.~\eqref{eq:att1}) with 
either the weighted sum (see Eq.~\eqref{eq:att_wsum})
or hard decision (see Eq.~\eqref{eq:att_hard} was
tested by training a model with the maximum likelihood estimator from
Sec.~\ref{sec:mle} and the variational learning from Sec.~\ref{sec:vari},
respectively. The authors of \cite{Xu-et-al-ICML2015} reported the similar
performances with these two approaches on a number of benchmark datasets.

\begin{table}[ht]
    \centering
    \caption{The performances of the image caption generation models in the
        Microsoft COCO Image Captioning Challenge. ($\star$)
        \cite{Vinyals-et-al-CVPR2015}, ($\bullet$) \cite{Fang-et-al-CVPR2015},
        ($\circ$) \cite{Devlin2015}, ($\diamond$) \cite{Donahue-et-al-arxiv2014} and
        ($\ast$) \cite{Xu-et-al-ICML2015}. The rows are sorted according to M1.}
    \label{tbl:capgen}
    \begin{tabular}{c|c c|c c}
        & \multicolumn{2}{c|}{Human} & \multicolumn{2}{c}{Automatic} \\
        Model & M1 & M2 & BLEU & CIDEr \\
        \hline
        \hline
        Human & 0.638 & 0.675 & 0.471 & 0.91\\
        Google$^\star$ & 0.273 & 0.317 & 0.587 & 0.946 \\
        MSR$^\bullet$ & 0.268 & 0.322 & 0.567 & 0.925 \\
        Attention-based$^\ast$ & 0.262 & 0.272 & 0.523 & 0.878 \\
        Captivator$^\circ$ & 0.250 & 0.301 & 0.601 & 0.937 \\
        Berkeley LRCN$^\diamond$ & 0.246 & 0.268 & 0.534 & 0.891 \\
    \end{tabular}
\end{table}

\subsubsection{Experimental Result}

In \cite{Xu-et-al-ICML2015}, the attention-based image caption generator was
evaluated on three datasets; Flickr 8K~\cite{Hodosh-et-al-JAIR2013}, Flickr
30K~\cite{Young-et-al-TACL2014} and MS CoCo~\cite{Lin-et-al-ECCV2014}. In
addition to the self-evaluation, an ensemble of multiple attention-based models
was submitted to Microsoft COCO Image Captioning
Challenge\footnote{\url{https://www.codalab.org/competitions/3221}} and
evaluated with multiple automatic evaluation metrics\footnote{
    BLEU~\cite{Papineni2002}, METEOR~\cite{Denkowski2014},
    ROUGE-L~\cite{Lin2004} and CIDEr~\cite{Vedantam2014}.
} as well as by human evaluators.

In this Challenge, the attention-based approach ranked third based on the
percentage of captions that are evaluated as better or equal to human caption
(M1) and the percentage of captions that pass the Turing Test (M2).
Interestingly, the same model was ranked eighth according to the most recently
proposed metric of CIDEr and ninth according to the most widely used metric of
BLEU.\footnote{\url{http://mscoco.org/dataset/\#leaderboard-cap}}
It means that this model has better relative performance in terms of human
evaluation than in terms of the automatic metrics, which only look at
matching subsequences of words, not directly at the meaning of the generated sentence.
The performance of the top-ranked systems, including the attention-based model
from \cite{Xu-et-al-ICML2015}, are listed in Table~\ref{tbl:capgen}. 

\begin{figure*}[!ht]           
    \centering         
    \includegraphics[width=0.87\textwidth]{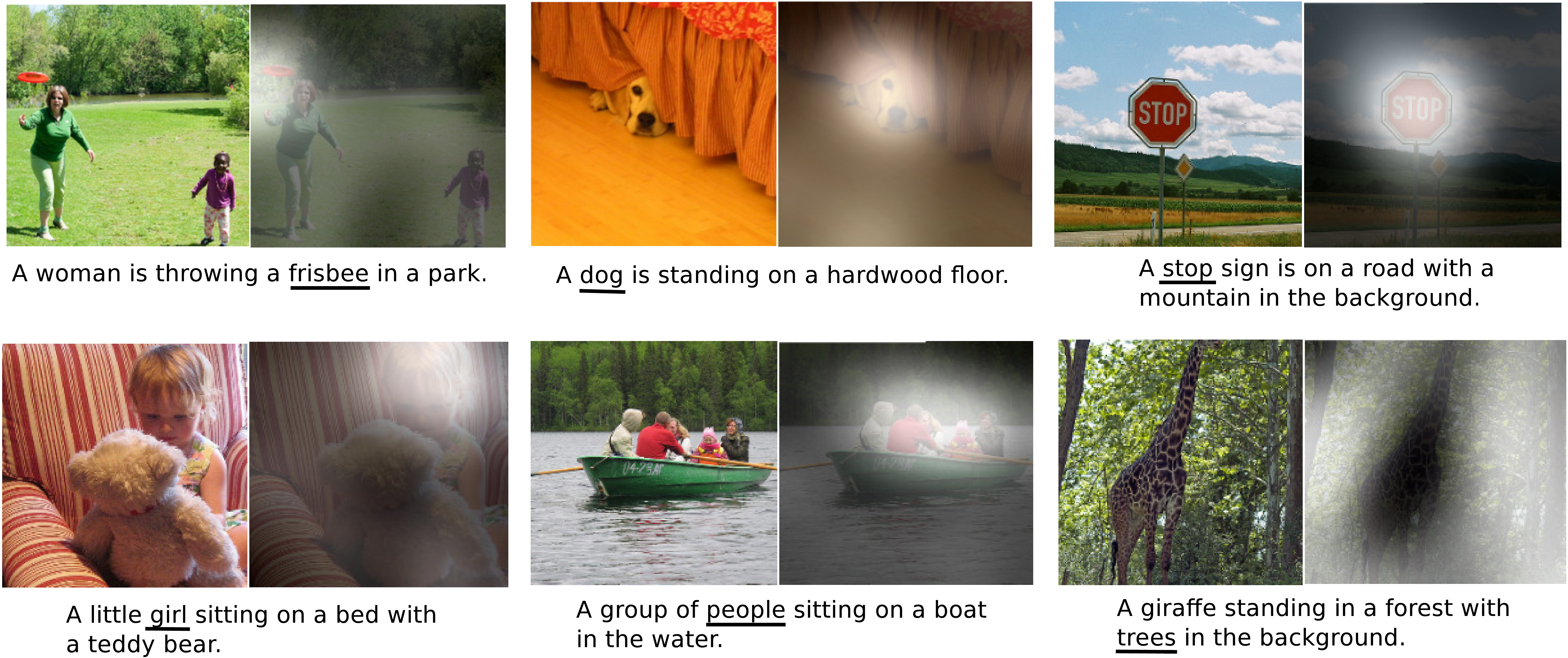}               
    \caption{Examples of the attention-based model attending to the correct
    object (\textit{white} indicates the attended regions, \textit{underlines}
indicated the corresponding word) \cite{Xu-et-al-ICML2015}}
    \label{fig:capgen_alignment}          
\end{figure*}

The attention-based model was further found to be highly interpretable,
especially, compared to the simple encoder--decoder models. See
Fig.~\ref{fig:capgen_alignment} for some examples.

\subsection{Video Description Generation}
\label{sec:capgen_vid}

Soon after the neural machine translation based on the simple encoder--decoder
framework was proposed in \cite{Sutskever-et-al-NIPS2014,Cho-et-al-EMNLP2014},
it was further applied to video description generation, which amounts to
translating a (short) video clip to its natural language description
\cite{Venugopalan-et-al-arxiv2014}. The authors of
\cite{Venugopalan-et-al-arxiv2014} used a pre-trained convolutional network
(see Sec.~\ref{sec:imagenet}) to extract a feature vector from each frame of the
video clip and average all the frame-specific vectors to obtain a single
fixed-dimensional context vector of the whole video. A conditional RNN-LM from
Sec.~\ref{sec:rnnlm} was used to generate a description based on this context
vector.

Since any video clip clearly has both temporal and spatial structures, it is
possible to exploit them by using the attention mechanism described throughout
this paper. In \cite{Li-et-al-ARXIV2015}, the authors proposed an approach based
on the attention mechanism to exploit the global and local temporal structures
of the video clips. Here we briefly describe their approach.

\begin{figure}[ht]
    \centering
    \includegraphics[width=\columnwidth,clip=True,trim=0 25 0 20]{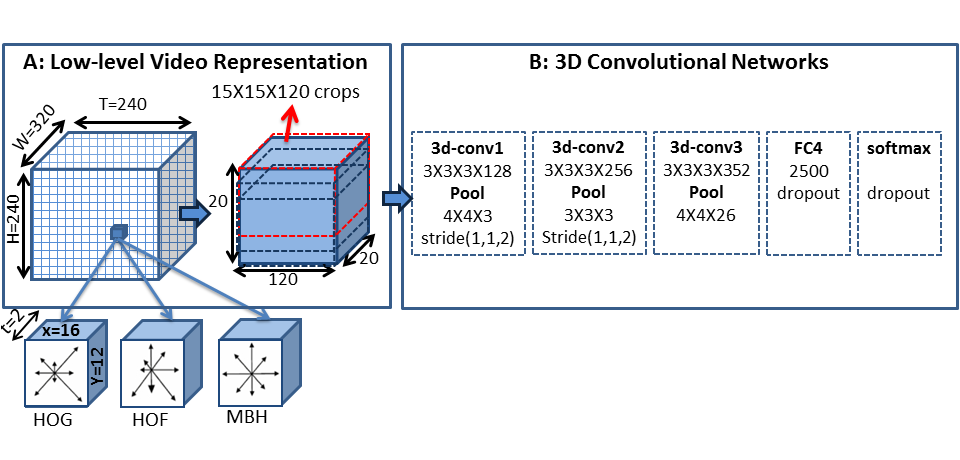}
    \caption{The 3-D convolutional network for motion from
    \cite{Li-et-al-ARXIV2015}.}
    \label{fig:3dconv}
\end{figure}

\subsubsection{Model Description}

In \cite{Li-et-al-ARXIV2015}, two different types of encoders are tested. The
first one is a simple frame-wise application of the pre-trained convolutional
network. However, they did not pool those per-frame context vectors as was done
in \cite{Venugopalan-et-al-arxiv2014}, but simply form a context set consisting
of all the per-frame feature vectors. The attention mechanism will work to
select one of those per-frame vectors for each output symbol being decoded. In
this way, the authors claimed that the overall model captures the {\it global}
temporal structure (the structure across many frames, potentially across the
whole video clip.) 

The other type of encoder in \cite{Li-et-al-ARXIV2015} is a so-called 3-D
convolutional network, shown in Fig.~\ref{fig:3dconv}. Unlike the usual convolutional network which often works
only spatially over a two-dimensional image, the 3-D convolutional network
applies its (local) filters across the spatial dimensions as well as the
temporal dimensions. Furthermore, those filters work not on pixels but on local
motion statistics, enabling the model to concentrate on motion rather than
appearance. Similarly to the strategy from Sec.~\ref{sec:imagenet}, the model
was trained on larger video datasets to recognize an action from each video
clip, and the activation vectors from the last convolutional layer were used as
context. The authors of \cite{Li-et-al-ARXIV2015} suggest that this encoder
extracts more local temporal structures complementing the global structures
extracted from the frame-wise application of a 2-D convolutional network.

The same type of decoder, a conditional RNN-LM, used in \cite{Xu-et-al-ICML2015}
was used with the content-based attention mechanism in Eq.~\eqref{eq:att1}.

\subsubsection{Experimental Result}

In \cite{Li-et-al-ARXIV2015}, this approach to video description generation has
been tested on two datasets; (1) Youtube2Text~\cite{Chen2011} and (2) Montreal
DVS~\cite{Torabi2015}. They showed that it is beneficial to have both types of
encoders together in their attention-based encoder--decoder model, and that the
attention-based model outperforms the simple encoder--decoder model. See
Table~\ref{tbl:capgen_vid} for the summary of the evaluation. 

\begin{table}[ht]
    \centering
    \caption{The performance of the video description generation models on
    Youtube2Text and Montreal DVS. ($\star$)~Higher the better. ($\circ$)~Lower
the better.}
    \label{tbl:capgen_vid}
    \begin{tabular}{c|c c |c c}
        & \multicolumn{2}{c}{Youtube2Text} & \multicolumn{2}{c}{Montreal DVS} \\
        Model & METEOR$^\star$ & Perplexity$^\circ$ & METEOR & Perplexity \\
    \hline
    \hline
Enc-Dec & 0.2868 & 33.09       & 0.044  & 88.28       \\
+ 3-D CNN & 0.2832 & 33.42       & 0.051  & 84.41       \\
+ Per-frame CNN & 0.2900 & 27.89       & .040  & 66.63       \\
+ Both & 0.2960 & 27.55       & 0.057  & 65.44       \\
    \end{tabular}
\end{table}

\begin{figure}[ht]
\centering
\includegraphics[width=0.9\columnwidth]{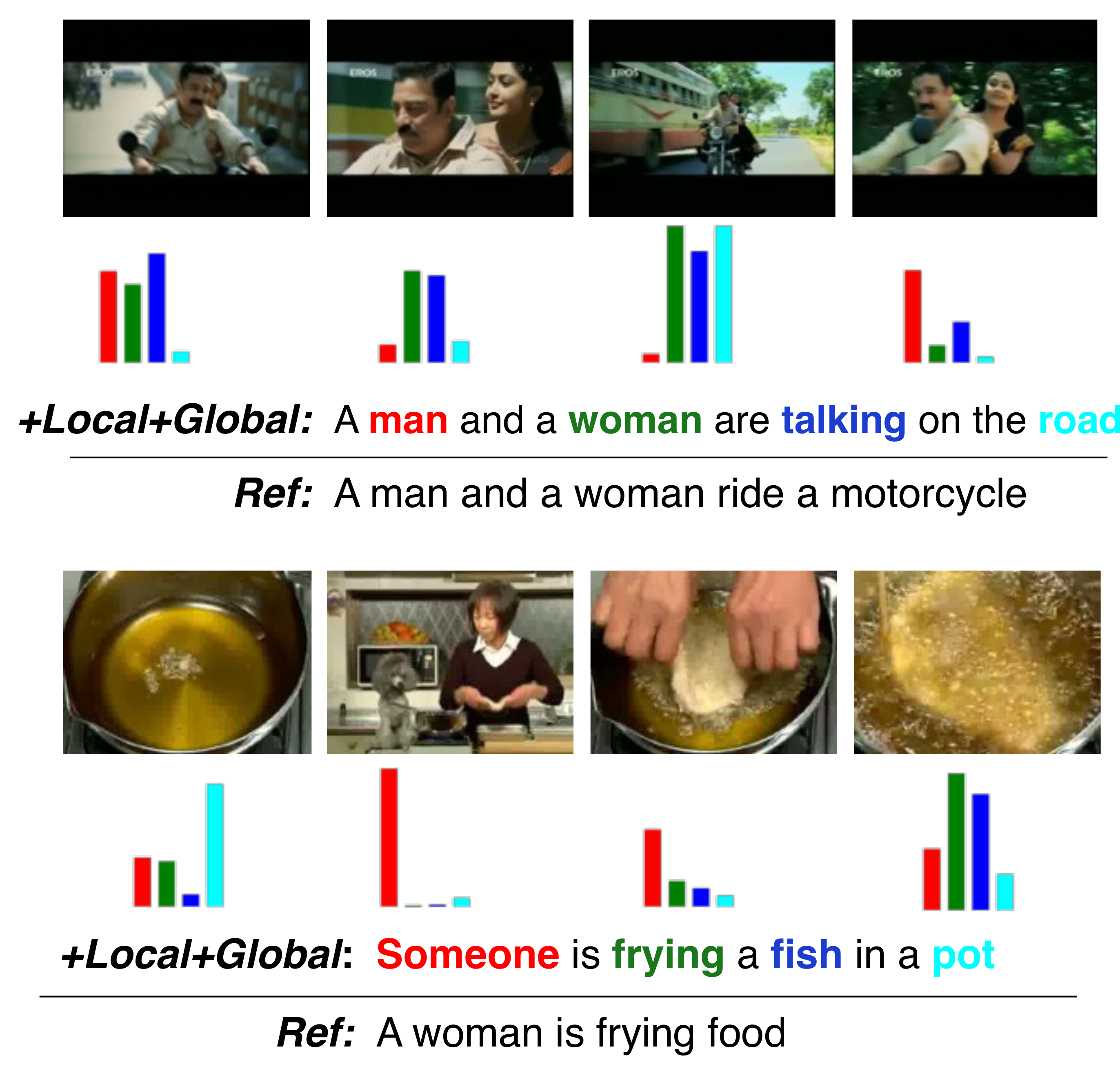}
\vspace{-2mm}
\caption{Two sample videos and their corresponding generated and ground-truth
    descriptions from Youtube2Text. The bar plot under each frame corresponds to
    the attention weight $\alpha_j^t$ (see
    Eq.~\eqref{eq:att2}) for the frame when the corresponding
    word (color-coded) was generated. Reprinted from
\cite{Li-et-al-ARXIV2015}.}
    \label{fig:capgen_vid_alignment}
\end{figure}

Similarly to all the other previous applications of the attention-based model,
the attention mechanism applied to the task of video description also
provides a straightforward way to inspect the inner
workings of the model. See
Fig.~\ref{fig:capgen_vid_alignment} for some examples. 

\subsection{End-to-End Neural Speech Recognition}
\label{sec:speech}

Speech recognition is a task in which a given speech waveform is translated
into a corresponding natural language transcription. Deep neural networks have
become a standard for the acoustic part of speech recognition systems
~\cite{deepSpeechReviewSPM2012}.
Once the input speech (often in the form of spectral
filter response) is processed with the deep neural network based acoustic
model, another model, almost always a hidden Markov model (HMM), is used to map
correctly the much longer sequence of speech into a shorter sequence of
phonemes/characters/words. Only recently, in
\cite{Graves-et-al-2006,Graves2012,Graves+Jaitly-ICML2014,Hannun2014}, fully neural
network based speech recognition models were proposed. 

Here, we describe the recently proposed attention-based fully neural speech
recognizer from \cite{Chorowski2015}. For more detailed comparison between the
attention-based fully speech recognizer and other neural speech recognizers,
e.g., from \cite{Graves+Jaitly-ICML2014}, we refer the reader to
\cite{Chorowski2015}. 

\subsubsection{Model Description--Hybrid Attention Mechanism}

The basic architecture of the attention-based model for speech recognition in
\cite{Chorowski2015} is similar to the other attention-based models described
earlier, especially the attention-based neural machine translation model in
Sec.~\ref{sec:nmt}. The encoder is a stacked bidirectional recurrent neural
network (BiRNN) \cite{Pascanu-et-al-ICLR2014} which reads the input sequence of
speech frames, where each frame is a 123-dimensional vector consisting of 40
Mel-scale filter-bank response, the energy and first- and second-order temporal
differences. The context set of the concatenated hidden states from the
top-level BiRNN is used by the decoder based on the conditional RNN-LM to
generate the corresponding transcription, which in the case of
\cite{Chorowski2015}, consists in a sequence of phonemes.

The authors of \cite{Chorowski2015} however noticed the peculiarity of speech
recognition compared to, for instance, machine translation. First, the lengths
of the input and output differ significantly; thousands of input speech frames
against a dozen of words. Second, the alignment between the symbols in the input
and output sequences is monotonic, where this is often not true in the case of
translation. 

These issues, especially the first one, make it difficult for the content-based
attention mechanism described in Eqs.~\eqref{eq:att1}~and~\eqref{eq:att2} to
work well. The authors of \cite{Chorowski2015} investigated these issues more
carefully and proposed that the attention mechanism with location awareness are
particulary appropriate (see Eq.~\eqref{eq:att_score_gen}. The location
awareness in this case means that the attention mechanism directly takes into
account the previous attention weights to compute the next ones.

The proposed location-aware attention mechanism scores each context vector by
\begin{align*}
    e_i^t = f_{\text{ATT}}(\vz_{t-1}, \vc_i, f_{\text{LOC}}^i(\left\{
        \alpha_j^{t-1}
    \right\}_{j=1}^T),
\end{align*}
where $f_{\text{LOC}}^j$ is a function that extracts information from the
previous attention weights $\left\{ \alpha_j^{t-1} \right\}$ for the $i$-th context
vector. In other words, the location-aware attention mechanism takes into
account both the content $\vc_i$ and the previous attention weights $\left\{
\alpha_j^{t-1} \right\}_{j=1}^T$. 

In \cite{Chorowski2015}, $f_{\text{LOC}}^j$ was implemented as
\begin{align}
    \label{eq:locatt2}
    f_{\text{LOC}}^j(\left\{ \alpha_j^t \right\}) =
    \sum_{k=j-\frac{K}{2}}^{j+\frac{K}{2}} \vv_k \alpha_{k}^{t-1},
\end{align}
where $K$ is the size of the window, and $\vv_k \in \RR^d$ is a
learned vector.

Furthermore, the authors of \cite{Chorowski2015} proposed additional
modifications to the attention mechanism, such as sharpening, windowing and
smoothing, which modify Eq.~\eqref{eq:att2}. For more details of each of these,
we refer the reader to \cite{Chorowski2015}.

\subsubsection{Experimental Result}

In \cite{Chorowski2015}, this attention-based speech recognizer was evaluated on
the widely-used TIMIT corpus~\cite{garofolo1993darpa}, closely following the
procedure from \cite{Graves-et-al-ICASSP2013}. As can be seen from
Table~\ref{tbl:speech}, the attention-based speech recognizer with the
location-aware attention mechanism can recognize a sequence of phonemes given a
speech segment can perform better than the conventional fully neural speech
recognition. Also, the location-aware attention mechanism helps the model
achieve better generalization error.

\begin{table}[ht]
    \caption{Phoneme error rates (PER). The bold-faced PER corresponds to the
    best error rate achieved with a fully neural network based model. From
\cite{Chorowski2015}.}
    \label{tbl:speech}
    \centering
    \begin{tabular}{l|c|c}
        \multicolumn{1}{c|}{\bf Model}  &\multicolumn{1}{c|}{\bf Dev} &\multicolumn{1}{c}{\bf Test} \\ 
        \hline 
        \hline 
        Attention-based Model & 15.9\% & 18.7\% \\
        Attention-based Model + Location-Awareness & 15.8\% & {\bf 17.6\%} \\
        \hline
        RNN Transducer~\cite{Graves-et-al-ICASSP2013} & N/A & 17.7\% \\
        \hline\hline
        Time/Frequency Convolutional Net+HMM
        \cite{Toth2014} & 13.9\% & 16.7\% 
    \end{tabular}
\end{table}

\begin{figure*}[ht]
  \centering
  \includegraphics[width=0.9\textwidth]{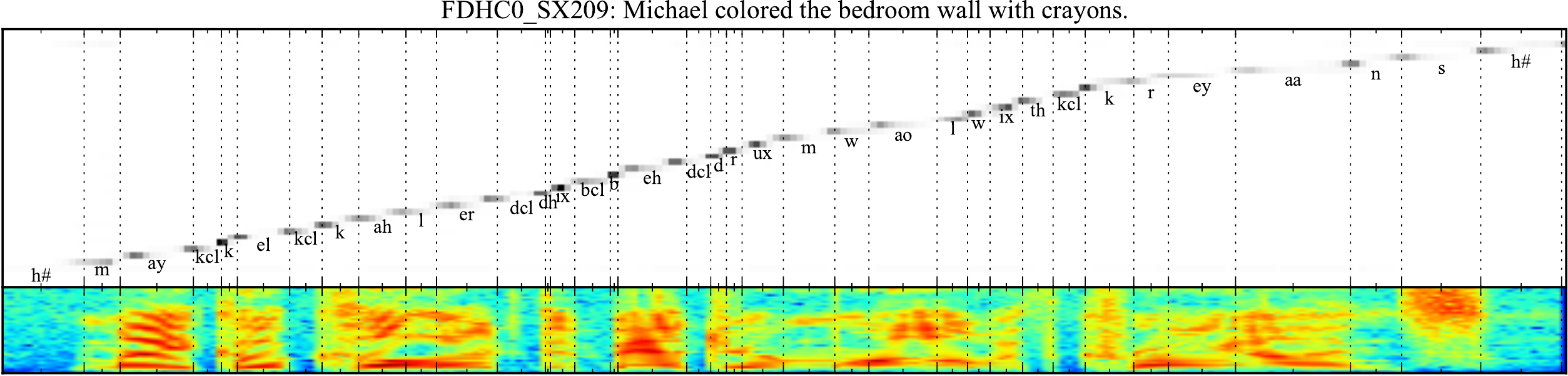}
  \caption{
      Attention weights by the attention-based model with
      location-aware attention mechanism.  The vertical
      bars indicate ground-truth phone location. For more details, see
      \cite{Chorowski2015}.
  }  
  \label{fig:speech}
\end{figure*}

Similarly to the previous applications, it is again possible to inspect the
model's behaviour by visualizing the attention weights. An example is shown in
Fig.~\ref{fig:speech}, where we can clearly see how the model attends to a
roughly correct window of speech each time it generates a phoneme.

\subsection{Beyond Multimedia Content Description}

We briefly present three recent works which applied the described
attention-based mechanism to tasks other than multimedia content description.

\subsubsection{Parsing--Grammar as a Foreign Language}

Parsing a sentence into a parse tree can be considered as a variant of machine
translation, where the target is not a sentence but its parse tree. In
\cite{Vinyals2014}, the authors evaluate the simple encoder--decoder model and
the attention-based model on generating the linearized parse tree associated with a natural
language sentence. Their experiments revealed that the attention-based parser
can match the existing state-of-the-art parsers which are often highly
domain-specific.

\subsubsection{Discrete Optimization--Pointer Network}

In \cite{Vinyals2015}, the attention mechanism was used to (approximately) solve
discrete optimization problems. Unlike the usual use of the described attention
mechanism where the decoder generates a sequence of output symbols, in their
application to discrete optimization, the decoder predicts which one of the
source symbols/nodes should be chosen at each time step.  The authors achieve
this by considering $\alpha_i^t$ as the probability of choosing the $i$-th input
symbol as the selected one, at each time step $t$. 

For instance, in the case of travelling salesperson problem (TSP), the model
needs to generate a sequence of cities/nodes that cover the whole set of input
cities so that the sequence will be the shortest possible route in the input map
(a graph of the cities) to cover every single city/node. First, the encoder
reads the graph of a TSP instance and returns a set of context vectors, each of
which corresponds to a city in the input graph.  The decoder then returns a
sequence of probabilities over the input cities, or equivalently the context
vectors, which are computed by the attention mechanism. The model is trained to
generate a sequence to cover all the cities by correctly {\it attending} to each
city using the attention mechanism. 

As was shown already in \cite{Vinyals2015}, this approach can be applied to any
discrete optimization problem whose solution is expressed as a subset of the
input symbols, such as sorting.

\subsubsection{Question Answering--Weakly Supervised Memory Network}
\label{sec:wsmn}

The authors of \cite{Sukhbaatar2015} applied the attention-based model to a
question-answering (QA) task. Each instance of this QA task consists of a set of
facts and a question, where each fact and the question are both natural language
sentences. Each fact is encoded into a continuous-space representation, forming a
context set of fact vectors. The attention mechanism is applied to the context
set given the continuous-space representation of the question so that the model
can focus on the relevant facts needed to answer the question. 

\section{Related Work: Attention-based Neural Networks}

The most related, relevant model is a neural network with {\it location-based}
attention mechanism, as opposed to the content-based attention mechanism
described in this paper. The content-based attention mechanism computes the
relevance of each spatial, temporal or spatio-temporally localized region of the
input, while the location-based one directly returns to which region the model
needs to attend, often in the form of the coordinate such as the $(x,y)$-coordinate
of an input image or the offset from the current coordinate.

In \cite{Graves-arxiv2013}, the location-based attention mechanism was
successfully used to model and generate handwritten text. In
\cite{ba+mnih,Mnih2014}, a neural network is designed to use the location-based
attention mechanism to recognize objects in an image. Furthermore, a generative
model of images was proposed in \cite{Gregor2015}, which iteratively reads and
writes portions of the whole image using the location-based attention mechanism.
Earlier works on utilizing the attention mechanism, both content-based and
location-based, for object recognition/tracking can be found in
\cite{Larochelle2010,Denil2012,Zheng2014}.

The attention-based mechanim described in this paper, or its variant, may be
applied to something other than multimedia input. For instance, in
\cite{Graves2014}, a neural Turing machine was proposed, which implements a
memory controller using both the content-based and location-based attention
mechanisms. Similarly, the authors of \cite{Weston2014} used the content-based
attention mechanism with hard decision (see, e.g., Eq.~\eqref{eq:att_hard}) to
find relevant memory contents, which was futher extended to the weakly
supervised memory network in \cite{Sukhbaatar2015} in Sec.~\ref{sec:wsmn}. 

\section{Looking Ahead...}

In this paper, we described the recently proposed attention-based
encoder--decoder architecture for describing multimedia content. We started by
providing background materials on recurrent neural networks (RNN) and
convolutional networks (CNN) which form the building blocks of the
encoder--decoder architecture. We emphasized the specific variants of those
networks that are often used in the encoder--decoder model; a conditional
language model based on RNNs (a conditional RNN-LM) and a pre-trained CNN for
transfer learning. Then, we introduced the simple encoder--decoder model
followed by the attention mechanism, which together form the central topic of
this paper, the attention-based encoder--decoder model.

We presented four recent applications of the attention-based encoder--decoder
models; machine translation (Sec.~\ref{sec:nmt}), image caption generation
(Sec.~\ref{sec:capgen}), video description generation (Sec.~\ref{sec:capgen_vid}) and
speech recognition (Sec.~\ref{sec:speech}). We gave a concise
description of the attention-based model for each of these applications together
with the model's performance on benchmark datasets. Furthermore, each
description was accompanied with a figure visualizing the behaviour of the
attention mechanism.

In the examples discussed above, the attention mechanism was primarily considered as a means to
building a model that can describe the input multimedia content in natural
language, meaning the ultimate goal of the attention mechanism was to aid the
encoder--decoder model for multimedia content description. However, this should
not be taken as the only possible application of the attention mechanism. 
Indeed, as recent work such as the pointer networks \cite{Vinyals2015} suggests, 
future applications of attention mechanisms could run the range of AI-related tasks. 

Beside superior performance it delivers, an attention mechanism can be used to extract
the underlying mapping between two entirely different modalities {\it without}
explicit supervision of the mapping.  From Figs.~\ref{fig:alpha_viz},
\ref{fig:capgen_alignment}, \ref{fig:capgen_vid_alignment} and \ref{fig:speech},
it is clear that the attention-based models were able to infer -- \emph{in an unsuperivsed way} -- alignments
between different modalities (multimedia and its text description) that
agree well with our intuition. This suggests that this type of attention-based
model can be used solely to extract these underlying, often complex, mappings
from a pair of modalities, where there is not much prior/domain knowledge. As an
example, attention-based models can be used in neuroscience to temporally
and spatially map between the neuronal activities and a sequence of
stimuli~\cite{Wehbe2014}.

%

\section*{Acknowledgment}

The authors would like to thank the following for research funding and
computing support: NSERC, FRQNT, Calcul Qu\'{e}bec, Compute Canada, the Canada Research
Chairs, CIFAR and Samsung.

\ifCLASSOPTIONcaptionsoff
  \newpage
\fi



\bibliographystyle{IEEEtran}
\bibliography{ml,aigaion}
\end{document}